\documentclass[conference]{IEEEtran}
\IEEEoverridecommandlockouts
\usepackage{cite}
\usepackage{amsmath,amssymb,amsfonts}
\usepackage{algorithmic}
\usepackage{graphicx}
\usepackage{textcomp}
\usepackage{xcolor}
\usepackage{hyperref}
\usepackage{url}
\usepackage{placeins}
\def\BibTeX{{\rm B\kern-.05em{\sc i\kern-.025em b}\kern-.08em
    T\kern-.1667em\lower.7ex\hbox{E}\kern-.125emX}}
\begin{document}

\makeatletter
\newcommand{\linebreakand}{%
  \end{@IEEEauthorhalign}% end current alignment row
  \hfill\mbox{}\par
  \mbox{}\hfill\begin{@IEEEauthorhalign}% start a new row
}
\makeatother

\title{Crop Recommendation and Agricultural Query Answering System
Using Spatio-Temporal Graph Neural Networks and Hybrid Retrieval
Augmentation}

\author{%
  % First row, two authors
  \IEEEauthorblockN{Prajwal Thapa}
    \IEEEauthorblockA{\textit{Kathmandu University}\\
    Dhulikhel, Nepal\\
    prazzwalthapa87@gmail.com}
  \and
  \IEEEauthorblockN{Yagya Raj Pandyea}
  \IEEEauthorblockA{\textit{Kathmandu University}\\
    Dhulikhel, Nepal\\
    yagya.pandeya@ku.edu.np}
}

\maketitle
\begin{abstract}
This paper presents a unified system designed to support precision agriculture by integrating advanced weather prediction, crop recommendation, and a question-answering tool for farmers. We propose two deep learning models—a Transformer-based Graph Neural Network and a Spatio-Temporal Graph Convolutional Network (STGCN)—to forecast weather conditions for the next 30 days using data from 1,359 locations in Nepal. The STGCN outperforms the Transformer-based model in accuracy (MSE $\approx$ 0.011 vs. 0.013), effectively modeling both spatial and temporal dependencies in climate data. These predictions are combined with static soil properties such as pH, moisture, and organic content to generate localized crop recommendations through a scoring algorithm that matches each crop's optimal growing conditions. Additionally, we develop a Retrieval-Augmented Generation (RAG) chatbot that leverages domain-specific agricultural documents to answer farmers’ questions in natural language. The entire system is deployed via a mobile application, offering real-time suggestions and conversational support. User feedback confirms the system’s usability and relevance, especially in rural settings where personalized farming guidance is limited. Overall, our approach demonstrates how combining machine learning models with local agricultural data can empower farmers with actionable insights, promoting more informed decisions, better crop yields, and increased resilience to climate variability.
\end{abstract}


\section{Introduction}
In Nepal, approximately 66\% of the population is involved in agriculture, which accounts for about a third of the country's GDP and plays a crucial role in the national economy. This sector provides a wide range of job opportunities, from farming to small-scale businesses. However, Nepal faces challenges such as food insecurity and frequent natural disasters, including floods, droughts, landslides, earthquakes, diseases, and pest outbreaks. The agriculture sector is predominantly characterized by subsistence farming, leading to low productivity and output of agricultural goods.

Traditionally, Nepalese farmers have relied on experience, historical weather patterns, and basic meteorological forecasts to make crop-related decisions. However, these methods often lack precision and fail to account for rapid climate changes and extreme weather events that have become more frequent in recent years due to global warming. The need for more sophisticated, data-driven approaches to agricultural decision-making has led to the development of advanced technologies, such as machine learning (ML) and artificial intelligence (AI).

Machine learning techniques have been widely used in agriculture for tasks like yield prediction \cite{7951740}, pest detection, and disease diagnosis. Recently, the advent of deep learning, particularly graph-based neural network and transformer-based models, has opened new possibilities for weather prediction and crop recommendation systems. Graph Neural Networks (GNN) \cite{4700287} and Transformers \cite{vaswani2017attention}, initially developed for natural language processing tasks and molecular chemistry, molecular biology, pattern recognition, and data mining, have demonstrated exceptional capabilities in handling sequential data, making them suitable for time-series forecasting, including weather prediction.

The core idea behind using spatio-temporal graph and transformer-based graph models for weather prediction is their ability to capture long-range dependencies and complex patterns in large datasets. These models can analyze historical weather data, learn from it, and make accurate predictions about future weather conditions. By integrating these predictions with other relevant agricultural data, such as soil properties and crop characteristics, a comprehensive crop recommendation system can be developed.

Such a system can provide farmers with timely and precise recommendations on which crops to plant based on predicted weather conditions, optimizing yields and reducing losses. Moreover, it can help mitigate the impacts of adverse weather events, enhancing the resilience of agricultural systems to climate variability and change.

In addition to the recommendation system, we also developed a knowledge-based RAG system designed to handle queries from farmers regarding various agricultural information. This includes details about crops, pest control, ongoing subsidies, and other relevant topics. To enhance the system's performance, we employed a hybrid approach using SPLADE and dense-sparse vectors for querying. SPLADE is particularly effective in retrieving sparse representations, which can capture relevant information with minimal computational overhead, and sparse vectors allow for more efficient matching of queries to the knowledge base. Dense vectors help to capture the semantic meanings. This approach ensures that farmers receive accurate and timely information, empowering them to take more informed actions in their farming practices, ultimately contributing to better outcomes in agriculture.

This thesis explores the development of a crop recommendation system that leverages transformer-based models for weather prediction. By combining advanced machine learning techniques with agricultural data, this research aims to create a tool that supports farmers in making informed decisions, ultimately contributing to more sustainable and productive agricultural practices.

\section{Related Work}
This chapter provides a detailed overview of the system's functionality, outlining its core components and the methodologies adopted for weather prediction and crop recommendation. It includes a review of the current research landscape and existing studies in transformer-based weather forecasting and agricultural decision support systems. Furthermore, the chapter identifies and discusses the limitations and challenges present in current systems, highlighting areas ripe for improvement to better assist farmers in making informed decisions grounded in precise weather forecasts and integrated soil data.

The increasing emphasis on precision agriculture has fueled significant research efforts aimed at developing crop recommendation systems that effectively integrate diverse environmental and geographical factors. These systems have evolved considerably, transitioning from simpler rule-based approaches to more sophisticated machine learning and deep learning models capable of incorporating dynamic variables like weather conditions, soil properties, and spatial information. For a country like Nepal, characterized by its varied climatic zones and soil types across 1359 distinct locations, such advanced systems are indispensable for maximizing agricultural output.

\subsection*{Advances in Crop Recommendation Systems}
The progression in developing crop recommendation systems is marked by the increasing incorporation of multiple data sources, including weather patterns, soil attributes, and spatial dependencies, into predictive models. Early research often relied on single-variable models, whereas recent efforts focus on integrating diverse datasets to achieve more accurate and holistic predictions.

\textbf{Weather Prediction in Agriculture:}
Weather prediction holds a central position in crop recommendation, given that environmental parameters such as temperature, rainfall, and humidity directly influence crop growth and development. \cite{mehra2017weather} presented one of the initial attempts to integrate weather prediction into agricultural decision-making using a linear regression model, though this model faced difficulties in handling non-linear weather patterns. Subsequent, more advanced models, including those described by \cite{kumari2020climate}, utilized algorithms like Random Forest and Support Vector Machines (SVM) to account for climate variability, demonstrating improved accuracy in crop yield prediction. More recently, deep learning methodologies, particularly transformer models, have been adapted for weather forecasting. \cite{liu2020transformer} introduced a transformer-based model that achieved significantly enhanced prediction of long-term weather patterns, attributed to its strong capability to capture temporal dependencies over extended sequences.

\textbf{Soil Data and Crop Recommendation:}
The critical importance of soil data, encompassing factors like pH levels, moisture content, and nutrient availability, has been consistently highlighted across various crop recommendation studies. \cite{chaudhari2018soil} implemented a system that relied exclusively on soil attributes to recommend suitable crops, employing decision trees for soil type classification. However, this method was inherently limited by its inability to account for dynamic environmental factors such as fluctuating weather conditions. \cite{sharma2018hybrid} advanced the field by integrating both weather and soil data into a hybrid machine learning model, which resulted in improved accuracy, particularly beneficial for specific regions with limited data availability. Despite this progress, many existing models still do not dynamically integrate temporal changes in soil properties, which is a crucial aspect for providing accurate recommendations over time.

\textbf{Spatio-Temporal Data Integration:}
Spatio-temporal data plays a vital role in the accurate modeling of agricultural systems, especially within geographically diverse countries like Nepal. Spatial data pertains to the geographical characteristics of a given location, while temporal data tracks changes in conditions over time. \cite{li2021stgnn} developed Spatio-Temporal Graph Neural Networks (STGNNs) specifically to analyze spatial and temporal data concurrently, proving useful for predicting crop yields across different geographical regions. This approach effectively addressed the challenge of modeling dynamic environmental conditions by incorporating both temporal sequences and spatial relationships into the analysis. Similarly, \cite{xu2022stgcn} extended this work, applying STGCNs to predict environmental conditions directly relevant to agriculture, further emphasizing the significance of spatio-temporal modeling in this domain. Nevertheless, the specific application of these advanced techniques directly for crop recommendation in agriculture remains an area with limited exploration.

\textbf{Graph Neural Networks in Agriculture:}
Graph Neural Networks (GNNs) have emerged as powerful tools particularly well-suited for handling spatial dependencies inherent in agricultural data, especially in environments involving multiple interconnected locations. \cite{zhang2018gnn} successfully applied GNNs to predict crop yields across a network of locations, effectively capturing the spatial relationships and dependencies between different regions. While beneficial for modeling regions with similar soil and weather characteristics, this approach did not incorporate temporal changes, which are fundamental for accurate and dynamic recommendations. \cite{wang2023crop} addressed this limitation by integrating both spatial and temporal data into a GNN model, leading to significant improvements in the performance of crop recommendation systems across varying environmental zones.

\textbf{Transformers in Agricultural Systems:}
Originally developed for state-of-the-art natural language processing tasks, transformer models have been successfully adapted for time-series forecasting applications within the agricultural sector. The attention mechanism, a key innovation introduced by \cite{vaswani2017attention}, has made transformers exceptionally useful for predicting long-term weather patterns due to their capability to focus on the most relevant parts of a sequence while disregarding less important information. \cite{lin2021transformers} developed a transformer-based model specifically for predicting crop yields, demonstrating that the model could effectively process and learn from large-scale agricultural datasets. This work underscored the potential for integrating transformers into agricultural systems, although further research is needed to combine the strengths of transformers with GNNs to build more robust and comprehensive crop recommendation systems.

\subsection*{Summary of Related Work}
In Table \ref{tab:contributions}, a summary of the major contributions from related studies in crop recommendation systems is presented, highlighting their various data sources, methodologies, and identified limitations.

\begin{table*}[htbp]
    \centering
    \begin{tabular}{p{3cm}|p{3cm}|p{3cm}|p{4cm}}
    \hline
    \textbf{Author(s)} & \textbf{Data Sources} & \textbf{Methods Used} & \textbf{Limitations} \\
    \hline
    \cite{mehra2017weather} & Weather data (rainfall, temperature) & Linear Regression & Poor performance on non-linear data, no soil or spatial analysis. \\
    \hline
    \cite{kumari2020climate} & Climate data (rainfall, humidity) & Random Forest, SVM & Improved accuracy but no integration of soil or spatial data. \\
    \hline
    \cite{chaudhari2018soil} & Soil data (pH, nutrients) & Decision Tree & Limited to static soil data, no temporal or weather integration. \\
    \hline
    \cite{sharma2018hybrid} & Weather and soil data & Hybrid Model (Soil + Weather) & No spatio-temporal consideration, regional application only. \\
    \hline
    \cite{liu2020transformer} & Weather time-series data & Transformer for time-series forecasting & Focused solely on weather data, without spatial or soil integration. \\
    \hline
    \cite{li2021stgnn} & Spatio-temporal weather and yield data & Spatio-Temporal Graph Neural Networks (STGNN) & Limited application in agriculture, focused on environmental data. \\
    \hline
    \cite{zhang2018gnn} & Spatial yield and soil data & Graph Neural Networks (GNNs) & Lacked temporal analysis, focusing only on spatial dependencies. \\
    \hline
    \cite{wang2023crop} & Weather, soil, and temporal data & GNN integrated with environmental data & Did not include transformer-based prediction for long-term trends. \\
    \hline
    \cite{lin2021transformers} & Weather data for crop yield & Transformer-based prediction & Only integrated weather data, no spatial or soil inclusion. \\
    \hline
    \cite{xu2022stgcn} & Spatio-temporal environmental data & Spatio-Temporal Graph Convolutional Networks & Focused on environmental factors, limited crop recommendation use. \\
    \hline
    \end{tabular}
    \vspace{0.2cm}
    \caption{Summary of Contributions in Crop Recommendation Systems}
    \label{tab:contributions}
\end{table*}

\subsection*{Research Gap}
Despite the considerable advancements made in developing crop recommendation systems, several significant gaps remain unaddressed in the existing body of research. Firstly, a common limitation is that most systems tend to focus primarily on either weather prediction or the analysis of soil data in isolation, failing to integrate both crucial data sources within a unified modeling framework that effectively accounts for spatio-temporal dependencies. While a few studies, notably \cite{li2021stgnn} and \cite{wang2023crop}, have successfully combined spatio-temporal data analysis with GNNs, these specific applications have not yet been adapted or tailored for direct use in agricultural crop recommendation systems. Furthermore, while transformer models have clearly demonstrated their effectiveness in time-series forecasting, particularly for weather prediction tasks, their integration with GNNs to simultaneously handle both spatial relationships and temporal dependencies within the context of agricultural crop recommendation systems remains an area that is significantly underexplored. Nepal's unique and highly diverse geographical landscape, encompassing data from 1359 distinct locations, presents a specific challenge for traditional models, as they often struggle to effectively account for both localized microclimates and broader regional or global trends in weather and soil conditions concurrently. This research is specifically designed to bridge these identified gaps by proposing and implementing a novel system that integrates the capabilities of transformer models for accurate weather prediction with Spatio-Temporal Graph Convolutional Networks (ST-GCNs) for comprehensive soil and spatial data analysis, a system specifically tailored to meet the distinct needs and challenges of Nepalese agriculture.

\section{Methodology}
This research employs advanced machine learning techniques for weather prediction and crop recommendation. The methodology consists of two main parts: weather prediction using Graph Neural Networks (GNNs) and Transformer-based Graph models, and crop recommendation based on predicted weather and soil data.

\subsection{Datasets}
Four datasets were utilized: location, weather, soil, and crop data.

\subsubsection{Location Data}
Location data for 1359 agricultural regions in Nepal was collected manually using \cite{googleEarth2024}. This dataset \cite{location_data} includes Longitude, Latitude, Altitude, and Location Name for diverse farming zones (see \ref{fig:location_data}).

\begin{figure*}[htbp]
  \centering
  \includegraphics[width=0.8\textwidth]{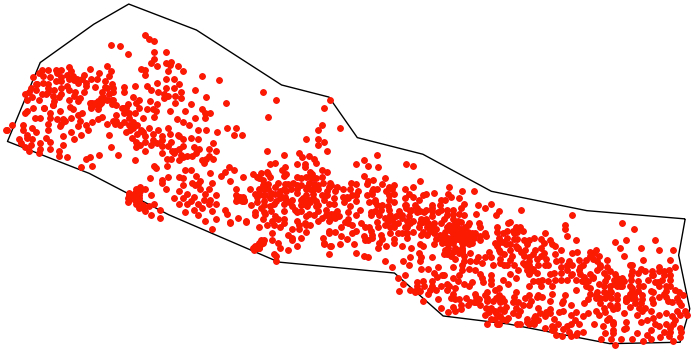}
  \caption{Locations selected all over Nepal}
  \label{fig:location_data}
\end{figure*}

\subsubsection{Weather Data}
A comprehensive weather dataset \cite{weather_data} covering 1359 locations over 42 years was obtained from NASA's \cite{powerLarc2024}. Key variables include Temperature (T2M), Precipitation (PRECTOTCORR), Humidity (RH2M), Wet Bulb Temperature (T2MWET), Wind speed (WS10M), and Earth Skin Temperature (TS).

Preprocessing involved correlation testing to select the top 6 features most relevant to Temperature, Humidity, and Precipitation. Data was then standardized using the formula:
\[
z_i = \frac{x_i - \mu_i}{\sigma_i}
\]
where \(x_i\) is the original value, \(\mu_i\) is the mean, and \(\sigma_i\) is the standard deviation of the feature. This ensures features contribute equally and improves model performance.

Temporal snapshots were created with a 120-day lag to capture seasonal variability. Locations were represented as nodes in a graph structure. Edges were defined based on either geodesic distance less than 15 km (using the Haversine formula) or altitude difference less than 50 meters. Edge weights were inversely proportional to the distance or altitude difference, capturing spatial and altitudinal relationships.

\subsubsection{Soil Dataset}
Soil data for the 1359 locations was collected via an API provided by \cite{soilNarc2024}. This dataset \cite{soil_data} includes Longitude, Latitude, Parent Soil, pH, Clay, Organic Matter, Total Nitrogen, and P2O5. Feature selection removed attributes less correlated with crop growth (e.g., sand, boron).

\subsubsection{Vegetable Dataset}
Vegetable data was curated from sources like \cite{comprehensive_insights_vegetables} and \cite{dry_fruit_booklet}, detailing optimal growth conditions. The dataset \cite{vegetable_data} \cite{fruit_data} includes Name, and minimum/maximum values for Soil pH, Phosphorus, Potassium, Nitrogen, Weather Temperature, Precipitation, Humidity, and Altitude. Feature selection focused on key growth factors.

\subsection{Weather Prediction}
Weather prediction utilizes historical data from the 1359 locations over 40 years. Preprocessing involved handling missing values (interpolation, forward-filling) and normalizing data. Temporal snapshots (weekly/monthly) were created to capture spatio-temporal patterns.

Two models were employed:
\subsubsection{Spatio-temporal Graph Convolutional Network (GCN)}
This model (\ref{fig:graph_representation}) uses a graph where nodes are locations. Edges represent geographical proximity or weather similarity. GCN layers capture spatial dependencies by propagating information between neighboring nodes. The model is extended to process sequences of graphs, capturing temporal trends.

\begin{figure*}[htbp]
    \centering
    \includegraphics[width=0.7\textwidth]{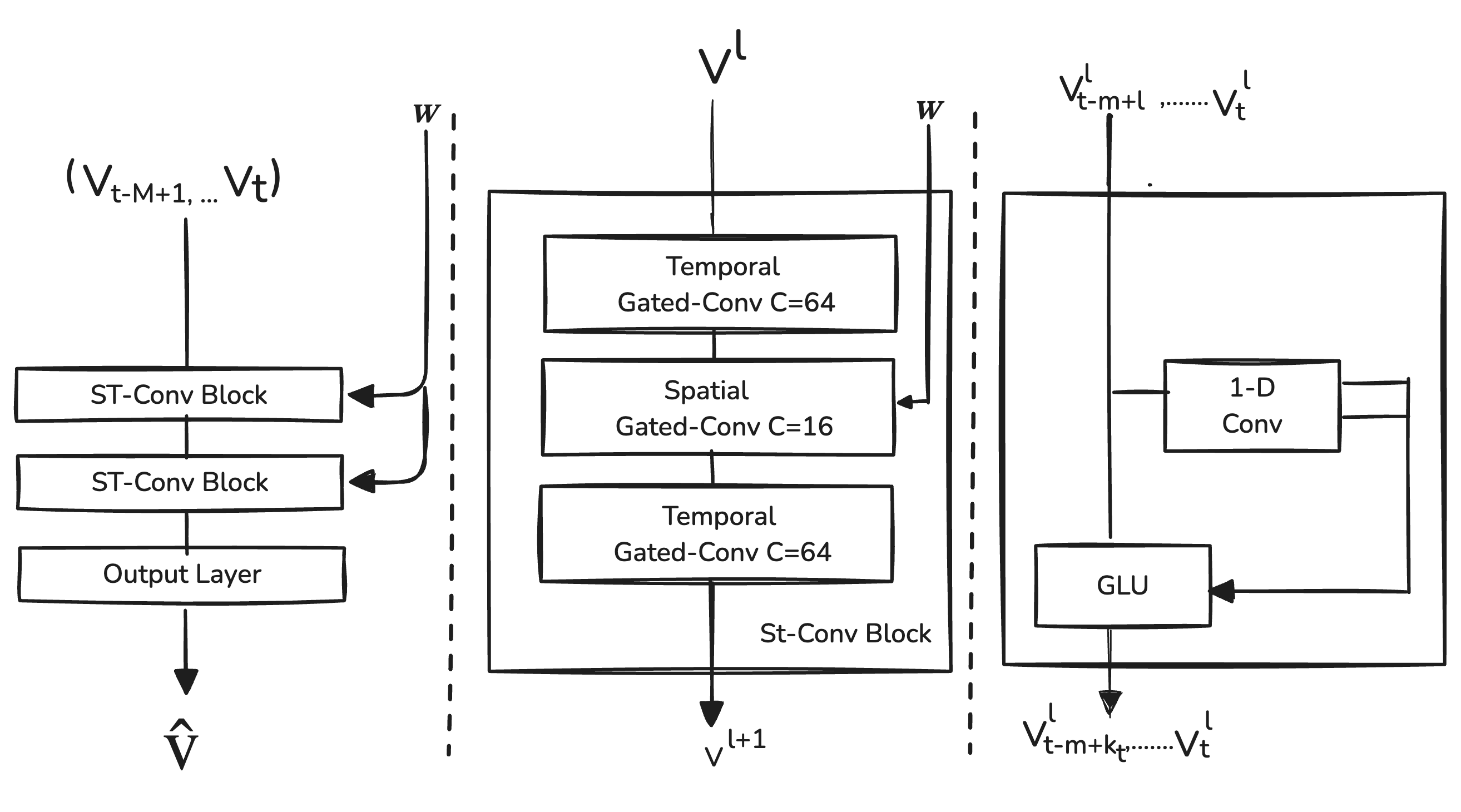}
    \caption{Architecture of Spatio temporal Graph Neural Networks.}
    \label{fig:graph_representation}
\end{figure*}

\subsubsection{Transformer-Based Graph Model}
The Transformer-Based Spatio-Temporal Graph Convolutional Network predicts multi-step data over the network of nodes. It combines temporal convolution (short-term patterns), a temporal transformer block (long-term dependencies), and graph convolution (spatial dependencies) before a final prediction layer.

\subsubsection{Training and Validation}
Models were trained using supervised learning with an Adam optimizer and a learning rate scheduler \cite{kingma2014adam}. Data was split 70\% training, 15\% validation, and 15\% testing. Hyperparameter tuning and cross-validation were used, along with early stoppage and dropout for regularization. Performance was evaluated using Mean Absolute Error (MAE) and Root Mean Square Error (RMSE).

\subsection{Crop Recommendation}
Crop recommendation integrates predicted weather data with soil data and historical crop information (\ref{fig:inferencing_weather_model}).

\begin{figure*}[htbp]
    \centering
    \includegraphics[width=1\textwidth, height=0.2\textheight]{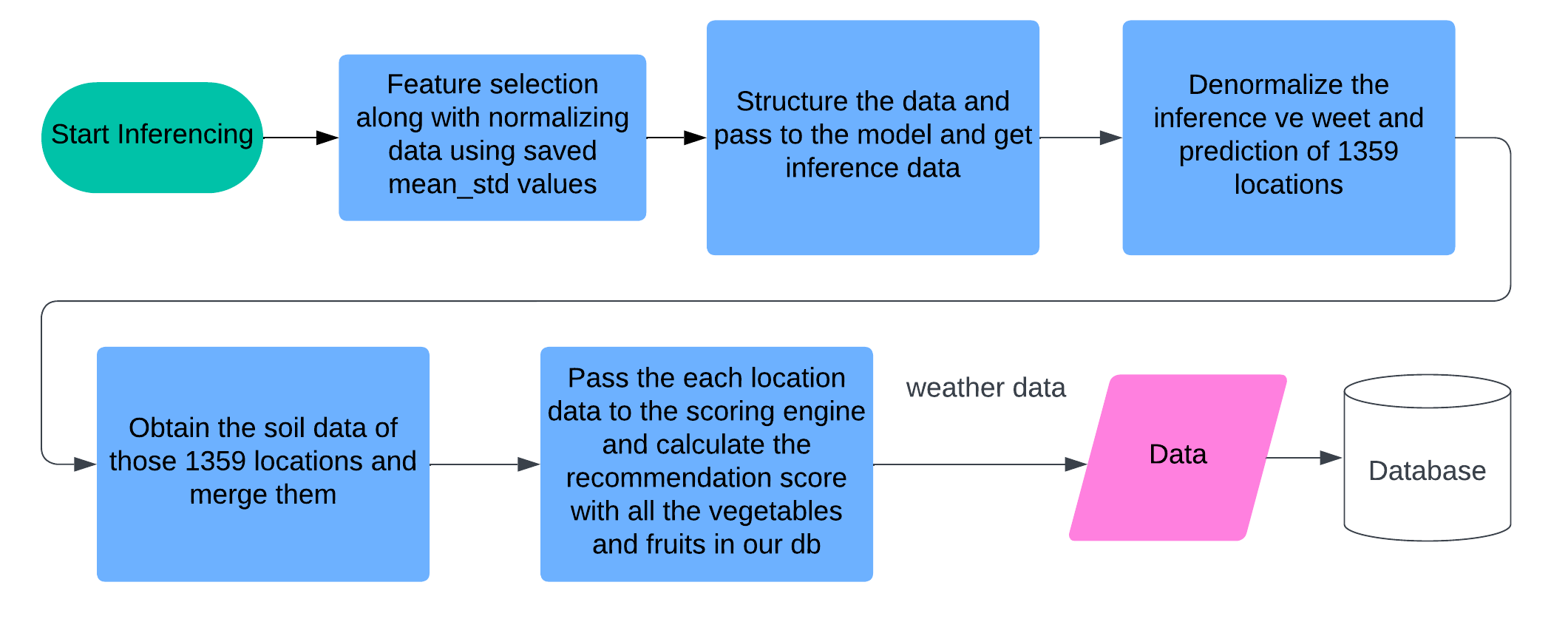}
    \caption{Flow Diagram of Crop Recommendation System (Model Inferencing)}
    \label{fig:inferencing_weather_model}
\end{figure*}

The recommendation engine matches location-specific environmental factors (predicted weather, soil data, altitude) with the optimal growth conditions listed in the vegetable dataset. A suitability score is calculated for each vegetable using an inverse distance metric:
\[
\text{score} += \frac{1}{1 + | \text{location\_data[feature]} - \text{mid\_val} |}
\]
where \(\text{mid\_val}\) is the midpoint of the vegetable's ideal range for a feature. Higher scores indicate better suitability. The engine outputs a ranked list of recommended crops for a given location.

\section{Experimental Setup}
The experiments were conducted at the A.I. Lab, Kathmandu University, located in Panchkhal, Kavre. The experimental setup utilized a local PC with specific hardware and software configurations. Despite facing challenges such as system shutdowns caused by overheating or insufficient memory resources, the experiments were successfully completed. Remote access to the PC was facilitated using an ngrok tunnel.

The hardware employed for training all models featured an 11th Gen Intel(R) Core(TM) i7-11700K CPU, an NVIDIA GeForce RTX 3080 Ti GPU with 12 GB GDDR6X memory and 10240 CUDA Cores operating at 1.67 GHz, and 16 GB (8GB x 2) of DIMM DDR4 Synchronous RAM running at 3200 MHz. The software environment was based on Ubuntu 22.04.3 LTS with kernel version 6.2.0-37-generic. Model implementations utilized Torch-geometric, built upon PyTorch 1.9.0+cu102, with Python 3.8.16 managed via Conda 23.1.0. The backend system was developed using Django and SQLite3, while the mobile application was built with the React Native framework. Pinecone was used as the cloud vector database for storing RAG documents, and Cohere's free API plan was utilized for retrieval augmentation.

\section{Results}

\subsection{Weather Prediction Model Performance}
To evaluate the prediction of 30-day weather parameters—humidity, precipitation, and temperature—we compared two models: a Spatio-Temporal Graph Neural Network (STGNN) and a transformer-based graph model. The key metric used was Mean Squared Error (MSE).

The STGNN achieved a lower MSE of 0.011 compared to 0.013 for the transformer-based model, indicating better performance in capturing spatial-temporal dependencies. This advantage is likely due to the STGNN’s ability to explicitly model interactions between geographical locations.

While the transformer-based model showed slightly higher MSE, it still performed well in capturing temporal patterns, thanks to its sequence modeling capabilities. However, its lack of spatial structure may limit its performance in spatially complex data.

\begin{table*}[htbp]
\centering
\begin{tabular}{c|c}
\hline
\textbf{Model} & \textbf{Mean Squared Error (MSE)} \\
\hline
Spatio-Temporal Graph Neural Network (STGNN) & 0.011 \\
Transformer-Based Graph Model & 0.013 \\
\hline
\end{tabular}

\vspace{0.2cm}  % add some space before caption
\caption{Comparison of Model Performance Based on MSE}
\label{tab:model_comparison}
\end{table*}

\begin{figure*}[htbp]
  \centering
  \includegraphics[width=0.8\textwidth]{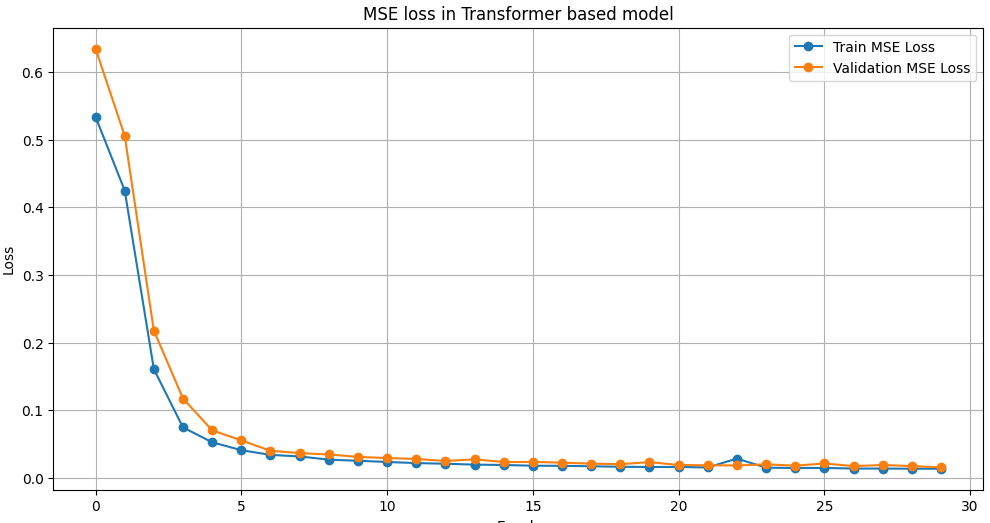}
  \caption{MSE loss for the Transformer Based Graph Model}
  \label{fig:transformer_mse}
\end{figure*}

\begin{figure*}[htbp]
  \centering
  \includegraphics[width=0.8\textwidth]{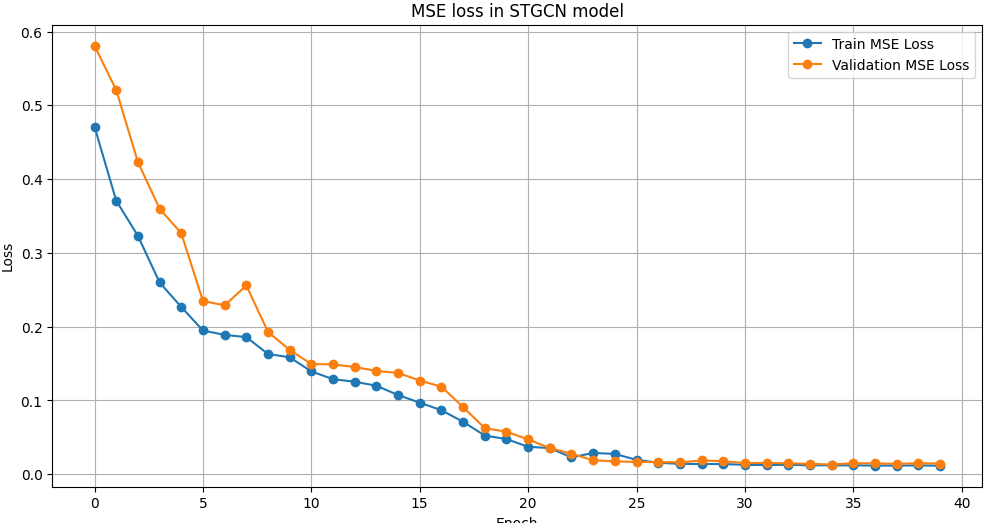}
  \caption{MSE loss for the STGCN}
  \label{fig:stgcn_loss}
\end{figure*}

\subsection{Mobile Interface}

The mobile application stands as a pivotal component of the vegetable and fruit recommendation system, functioning as the primary interface through which users, specifically farmers, interact with the system. By offering an intuitive and user-friendly platform, it successfully bridges the gap between sophisticated backend algorithms and the end-users, providing seamless, location-based recommendations directly at their fingertips. The application is developed with support for the Nepali language, which significantly enhances its usability and accessibility for local farmers. Key features integrated into the application include an easy-to-navigate interface, real-time updates on crop recommendations tailored to the user's location, and functionalities that could potentially allow for customization of recommendations based on various needs or preferences.

Upon launching the application, it initially requests location permission from the user. Once permission is granted, the system communicates with the backend to retrieve recommended crops specifically suited for the farmer's geographical location and current conditions. The application then displays information about the suitable crops, including details about their ideal growing conditions derived from the dataset, alongside environmental attributes predicted by our system, and general descriptive information about the crops themselves. This comprehensive presentation helps farmers understand the rationale behind the recommendations. Furthermore, the application extends its utility by providing information not only on the recommended crops but also on other vegetables and fruits that might be of interest to farmers. Users can browse a list of these crops and access detailed information about each one.This mobile platform offers unparalleled accessibility, enabling farmers to access valuable recommendations and information anytime, anywhere, directly supporting more informed decisions in their farming practices.

\subsection{RAG System for Farmer Queries}
A Retrieval-Augmented Generation (RAG) system was implemented as a crucial component to effectively address queries posed by farmers. This system is designed to combine the strengths of information retrieval with text generation capabilities. By first retrieving relevant information from a knowledge base and then utilizing a generation model to formulate context-sensitive answers, the system aims to provide clear, practical, and accurate solutions to common agricultural questions encountered by farmers.

The RAG system underwent testing with a diverse range of queries submitted by farmers, covering topics that included specific crop recommendations, methods for pest control, advice related to soil health management, and inquiries about ongoing agricultural subsidies. Observations from these tests indicated that the system consistently generated responses that were both relevant to the query and helpful to the farmer. The retrieval component played a key role in ensuring that the generated answers were well-grounded in accurate information obtained from the knowledge base. Simultaneously, the generation model effectively tailored the responses to directly address the specific needs and contexts expressed in each individual query. The system supports queries in multiple convenient languages, including Nepali and English, and is configured to provide the final answers in Nepali language, enhancing accessibility for local users. The RAG system employs a hybrid approach for document retrieval, combining methods like SPLADE with dense and sparse vector representations to improve the accuracy and efficiency of information retrieval from the knowledge base.

While quantitative metrics were not systematically applied for evaluation, qualitative assessments based on user interactions and feedback revealed that the RAG system provides valuable assistance to farmers.

\section{Conclusion}
This research developed a localized crop recommendation system for Nepal, leveraging weather data from 1359 locations, predicted using Spatio-Temporal Graph Neural Networks (STGNN) and Transformer-based graph models. Both models predicted 30-day weather parameters based on 100 days of historical data. The STGNN demonstrated superior performance with a Mean Squared Error (MSE) of 0.011, compared to the Transformer's MSE of 0.013, primarily due to its effectiveness in capturing spatio-temporal dependencies crucial for regional weather interactions. While the Transformer exhibited strength in long-term temporal patterns, the STGNN's graph-based approach proved more adept for this application.

The system integrates these weather predictions with local soil attributes using a scoring algorithm to provide tailored crop recommendations. Key contributions include the collection and preparation of localized weather and soil data for Nepal, the development and evaluation of STGNN and Transformer models for weather forecasting, the integration of weather and soil data for recommendations, and the implementation of a chat system to address farmer queries. This work provides a robust framework and highlights the potential of advanced AI models in agricultural planning.

\section{Limitations and Future Work}
Despite its promise, the system has limitations. It relies on potentially inconsistent historical data, lacks comprehensive quantitative evaluation metrics specifically for crop recommendation accuracy, and the models may not fully capture highly complex environmental interactions. Localization challenges regarding microclimates and cultural practices, and accessibility of necessary technology for all farmers, also exist.

Future work should focus on expanding data sources to include real-time satellite data and local reports, developing robust quantitative metrics for recommendation performance, enhancing model complexity with hybrid architectures, adapting recommendations for microclimates and local factors, and improving user accessibility through a mobile application with offline capabilities. Adapting the framework for other regions with similar agricultural contexts is also a valuable future direction.
%%%%%%%%%%%%%%%%%%%%%%%%%%%%%%%%%%%%%%%%%%%%%%%%%
\bibliographystyle{unsrt}
\bibliography{ref}
%\nocite{*}

\appendix
\section{Appendix}

\subsection{Detailed Data Variables}

\subsubsection*{Location Data Variables}
The variables collected for each location include geographical coordinates and elevation, crucial for mapping the diverse agricultural landscape of Nepal. The specific variables are detailed in Table \ref{table:location_data}.

\begin{table*}[htbp]
\centering
\begin{tabular}{c|c|c|c}
\hline
\textbf{Variable} & \textbf{Description} & \textbf{Unit} & \textbf{Data Type} \\ \hline
\textbf{Longitude} & Geographical east-west coordinate & Degrees & Float \\ \hline
\textbf{Latitude} & Geographical north-south coordinate & Degrees & Float \\ \hline
\textbf{Altitude} & Elevation above sea level & Meters & Float \\ \hline
\textbf{Location Name} & Name of the location & N/A & String \\ \hline
\end{tabular}
\vspace{0.2cm}
\caption{Data variables collected for each location.}
\label{table:location_data}
\end{table*}

\subsubsection*{Weather Data Variables}
The comprehensive weather dataset includes several key variables essential for forecasting and understanding their impact on crop growth. These variables are listed and described in Table \ref{table:weather_data}.

\begin{table*}[htbp]
\centering
\begin{tabular}{c|c|c|c}
\hline
\textbf{Variable} & \textbf{Description} & \textbf{Unit} & \textbf{Data Type} \\ \hline
\textbf{Temperature (T2M)} & Average daily temperature at 2m height & °C & Float \\ \hline
\textbf{Precipitation (PRECTOTCORR)} & Daily precipitation & mm & Float \\ \hline
\textbf{Humidity (RH2M)} & Daily average relative humidity at 2m height & \% & Float \\ \hline
\textbf{Wet Bulb Temperature (T2MWET)} & Wet bulb temperature at 2m height & °C & Float \\ \hline
\textbf{Wind Speed (WS10M)} & Wind speed at 10m height & m/s & Float \\ \hline
\textbf{Earth Skin Temperature (TS)} & Average daily earth surface temperature & °C & Float \\ \hline
\end{tabular}
\vspace{0.2cm}
\caption{Weather data variables used in the model.}
\label{table:weather_data}
\end{table*}

\subsubsection*{Soil Data Variables}
The soil dataset provides essential information on soil fertility and characteristics for each location, used in the crop recommendation process. The dataset contains the following attributes:

\begin{itemize}
  \item \textbf{Location}: The name or identifier of the location where the soil sample was collected.
  \item \textbf{Longitude}: The east-west geographic coordinate of the sampling point.
  \item \textbf{Latitude}: The north-south geographic coordinate of the sampling point.
  \item \textbf{Parent Soil}: The type of parent material or soil classification (e.g., alluvial, loamy).
  \item \textbf{pH}: The pH level of the soil, indicating its acidity or alkalinity.
  \item \textbf{Clay}: The percentage of clay particles in the soil.
  \item \textbf{Organic Matter}: The percentage of organic material in the soil, contributing to its fertility.
  \item \textbf{Total Nitrogen}: The total nitrogen content of the soil, a key nutrient for plant growth.
  \item \textbf{P2O5}: The phosphorus content measured as available phosphate (P2O5), crucial for root development.
\end{itemize}

\subsubsection*{Vegetable Dataset Variables}
The vegetable dataset outlines the optimal growth conditions for various crops, enabling tailored recommendations based on environmental and soil factors. The dataset includes the following attributes:

\begin{itemize}
  \item \textbf{Name}: The name of the vegetable.
  \item \textbf{Soil pH Min}: The minimum soil pH level suitable for the vegetable's growth.
  \item \textbf{Soil pH Max}: The maximum soil pH level suitable for the vegetable's growth.
  \item \textbf{Soil Phosphorus Min}: The minimum soil phosphorus level (as P2O5) required.
  \item \textbf{Soil Phosphorus Max}: The maximum phosphorus level (as P2O5) suitable.
  \item \textbf{Soil Potassium Min}: The minimum potassium level required (as K2O).
  \item \textbf{Soil Potassium Max}: The maximum potassium level suitable (as K2O).
  \item \textbf{Soil Nitrogen Min}: The minimum nitrogen level required.
  \item \textbf{Soil Nitrogen Max}: The maximum nitrogen level suitable.
  \item \textbf{Weather Temperature Min}: The minimum temperature for optimal growth.
  \item \textbf{Weather Temperature Max}: The maximum temperature suitable for the vegetable.
  \item \textbf{Weather Precipitation Min}: The minimum precipitation (mm) required.
  \item \textbf{Weather Precipitation Max}: The maximum precipitation the vegetable can tolerate.
  \item \textbf{Weather Humidity Min}: The minimum humidity level required.
  \item \textbf{Weather Humidity Max}: The maximum humidity level suitable.
  \item \textbf{Altitude Min}: The minimum altitude (in meters) where the vegetable can grow.
  \item \textbf{Altitude Max}: The maximum altitude where the vegetable can grow.
  \item \textbf{Matching Score}: A score representing how well the vegetable matches environmental and soil conditions.
\end{itemize}

\subsection{Dataset Availability}
The datasets utilized in this research are made available for further study and reproducibility. The location data is available at \cite{location_data}. The weather data, originally sourced from NASA Power \cite{powerLarc2024}, is provided at \cite{weather_data}. The soil data, obtained through the NARC API \cite{soilNarc2024}, can be accessed at \cite{soil_data}. The vegetable and fruit datasets, specifically curated for optimal crop growth conditions, are available at \cite{vegetable_data} and \cite{fruit_data}, respectively.

\subsection{Mobile Application Screenshots}
Screenshots illustrating additional details of the mobile application interface and its features are provided below, complementing the main description in the Results section.

\begin{figure}[htbp]
  \centering
  \includegraphics[width=0.85\columnwidth,height=15cm,keepaspectratio]{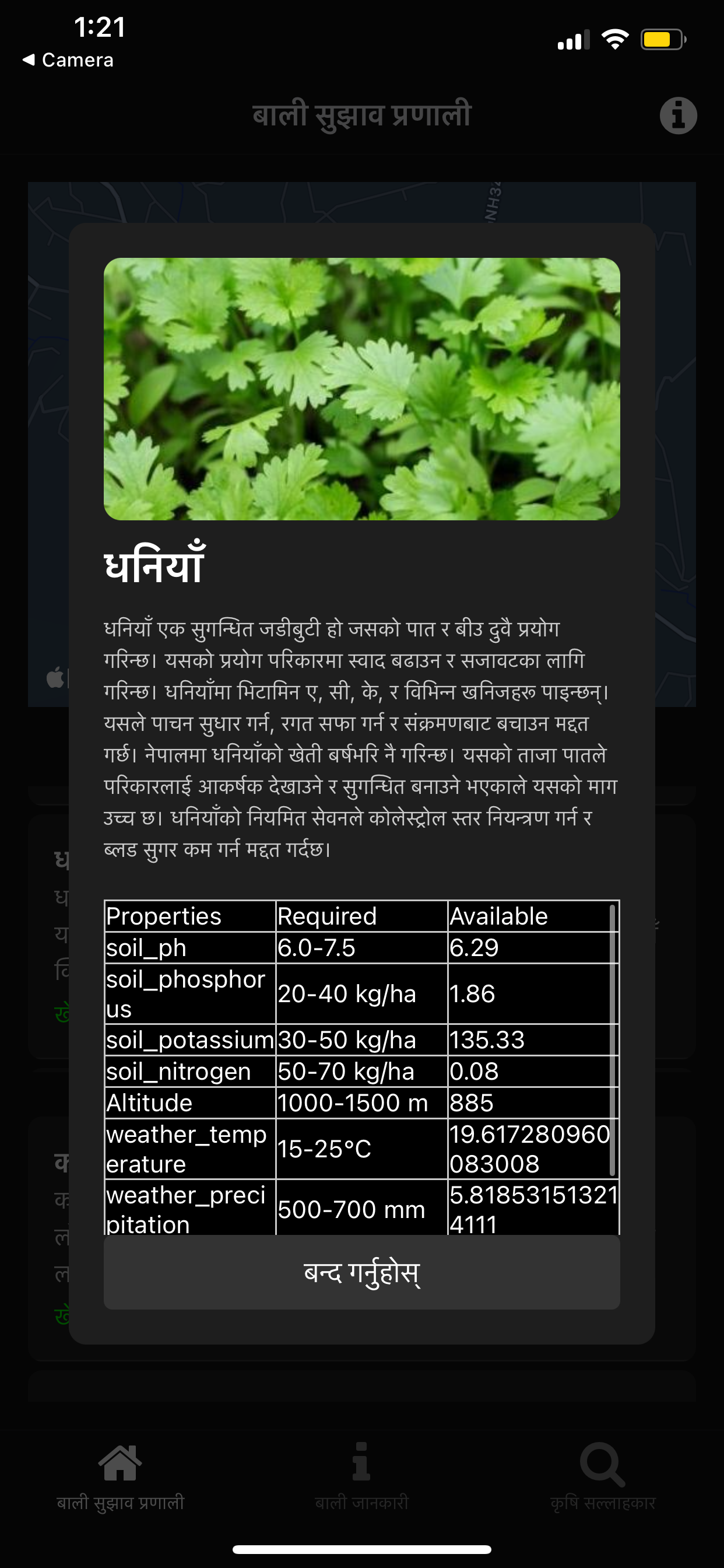}
  \caption{Details of recommended crop}
  \label{fig:mobile_application_crop_details}
\end{figure}

\begin{figure}[htbp]
  \centering
  \includegraphics[width=0.85\columnwidth,height=15cm,keepaspectratio]{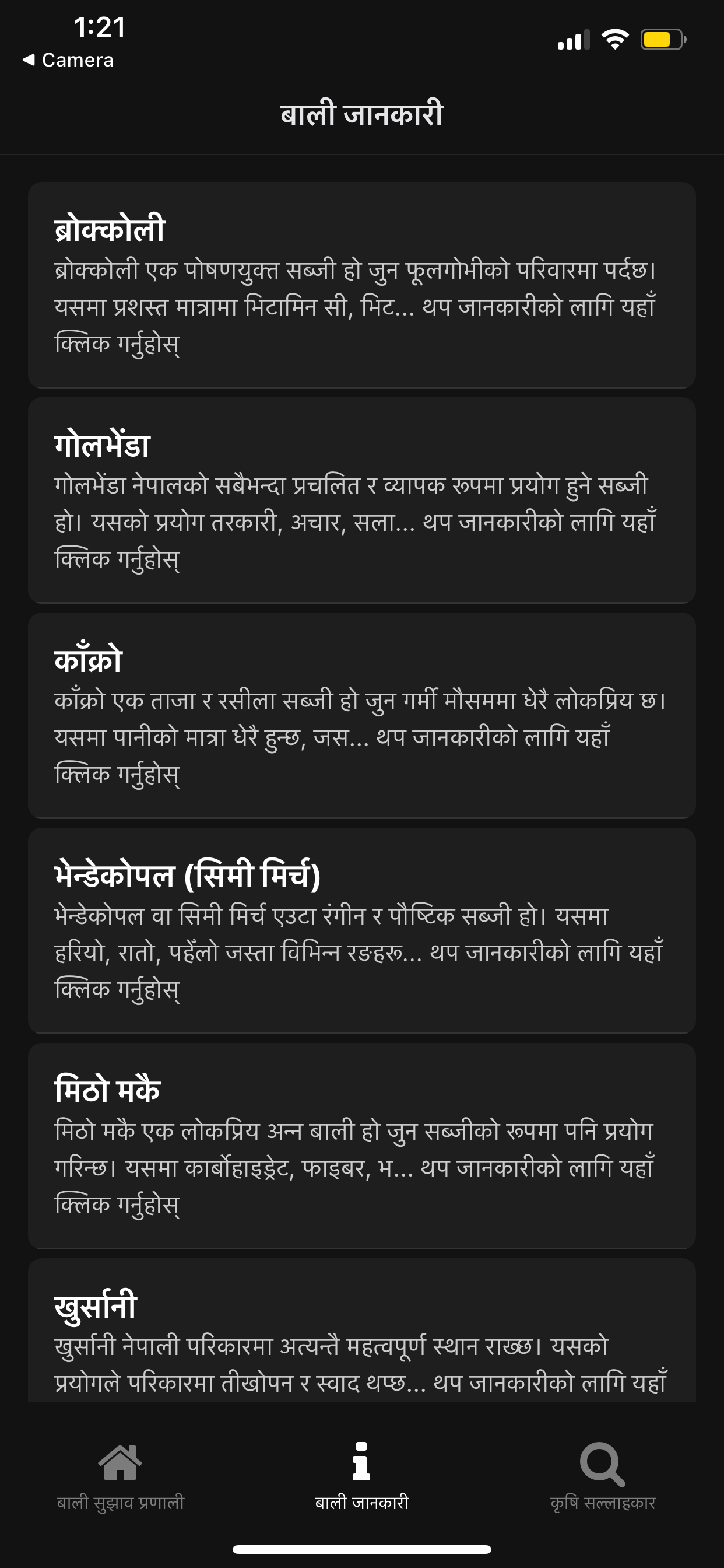}
  \caption{List of vegetables}
  \label{fig:mobile_application_vegetable_list}
\end{figure}

\begin{figure}[htbp]
  \centering
  \includegraphics[width=0.85\columnwidth,height=15cm,keepaspectratio]{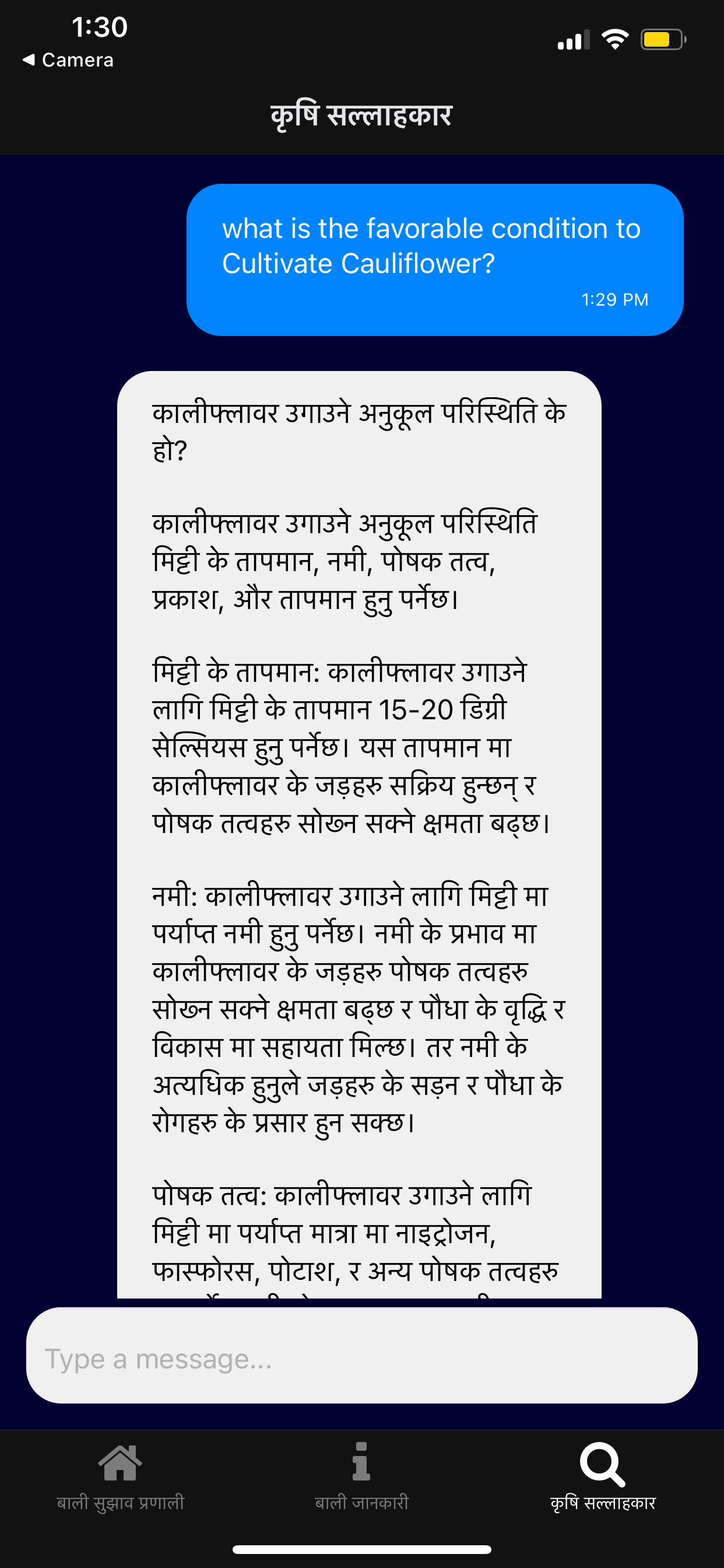}
  \caption{RAG implementation screenshot 1}
  \label{fig:rag_1}
\end{figure}
\end{document}